\documentclass[letterpaper, 10 pt, conference]{ieeeconf}  
\IEEEoverridecommandlockouts                              

\usepackage{graphicx}
\usepackage{amsmath}
\usepackage{amssymb}
\usepackage{booktabs}
\usepackage{subcaption}
\usepackage{comment}
\usepackage{svg}
\usepackage{cleveref}

\newcommand{\xtarget}{x_{\text{target}}}
\newcommand{\norm}[1]{\left\lVert#1\right\rVert}
\title{\LARGE \bf
Zero-Shot Object Searching \\Using Large-scale Object Relationship Prior
}

\author{Hongyi Chen$^{1}$, Ruinian Xu$^{1}$, Shuo Cheng$^{2}$, Patricio A. Vela$^{1}$, Danfei Xu $^{2}$ 
\thanks{$^{1}$H. Chen, R. Xu and P.A. Vela are with the School of Electrical and Computer Engineering, Georgia Institute of Technology, Atlanta, GA 30308, USA.
{\tt\small \{hchen657, pvela\}@gatech.edu}}%
\thanks{$^{2}$S. Cheng and D. Xu the School of Interactive Computing, Georgia Institute of Technology, Atlanta, GA 30308, USA.
{\tt\small \{shuocheng, danfei\}@gatech.edu}}%
}

\begin{document}

\maketitle
\thispagestyle{empty}
\pagestyle{empty}

\begin{abstract}

Home-assistant robots have been a long-standing research topic, and one of the biggest challenges is searching for required objects in housing environments. Previous object-goal navigation requires the robot to search for a target object category in an unexplored environment, which may not be suitable for home-assistant robots that typically have some level of semantic knowledge of the environment, such as the location of static furniture. In our approach, we leverage this knowledge and the fact that a target object may be located close to its related objects for efficient navigation. To achieve this, we train a graph neural network using the Visual Genome dataset to learn the object co-occurrence relationships and formulate the searching process as iteratively predicting the possible areas where the target object may be located. This approach is entirely zero-shot, meaning it doesn't require new accurate object correlation in the test environment. We empirically show that our method outperforms prior correlational object search algorithms. As our ultimate goal is to build fully autonomous assistant robots for everyday use, we further integrate the task planner for parsing natural language and generating task-completing plans with object navigation to execute human instructions. We demonstrate the effectiveness of our proposed pipeline in both the AI2-THOR simulator and a Stretch robot in a real-world environment.
\end{abstract}

\section{INTRODUCTION}


Home-assistant robots that share the same indoor space with humans are designed to assist them by interpreting human instructions and completing the corresponding home activities. In related work, researchers have evaluated the different aspects of home-assistant robots \cite{yamazaki2012home}\cite{SGL2022}\cite{tang2021learning}. Our paper focuses on developing an efficient object searching algorithm in the household environment for assistant robots.

Object search is a crucial ability for home-assistant robots and has been extensively studied \cite{batra2020objectnav}. Various approaches such as end-to-end reinforcement or imitation learning are commonly used to search for objects in unexplored environments \cite{mousavian2019visual}\cite{gupta2017cognitive}\cite{batra2020objectnav}. However, for home-assistant robots that are accustomed to their daily working environments, it is more practical to use partially observable Markov decision process (POMDP)-based approaches to maintain and update a measure of uncertainty over the location of the target object via its belief state, instead of building an episodic map \cite{8793494}\cite{li2016act}\cite{holzherr2021efficient}. Existing POMDP-based approaches typically make assumptions about object independence to reason about belief states, which makes them scalable but less accurate in their search for objects \cite{holzherr2021efficient}. Alternatively, these approaches may rely on accurate object correlation information, which may not always be available \cite{zheng2022towards}. In such cases, the success rate of the search process may decrease significantly.

\begin{figure}
     \centering
     \includegraphics[width=1\columnwidth]{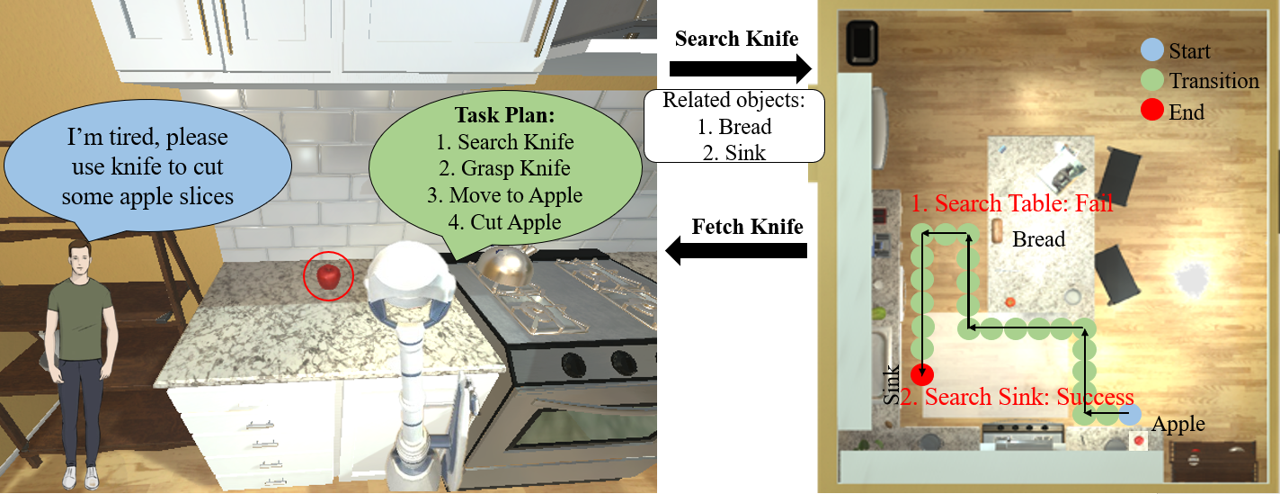}
     \caption{Illustration of the proposed framework for object searching and human instruction following. The human query is parsed into action sequence. Based on learned object co-occurrence relationships, GNN predicts possible areas to search for missing \texttt{knife} and complete the generated plan.}
     \label{fig:teaser}
\vspace{-10pt}
\end{figure}


This paper proposes a new approach for object search that does not rely on accurate object correlation information. Instead, the approach leverages the co-occurrence between object pairs to facilitate the search. Our approach involves predicting related objects whose positions are already known to the robot (e.g., furniture, fixtures) and searching areas with a high probability of finding a particular object. For instance, a knife is likely to be placed on a kitchen countertop. To model the object relationships, we construct a graph neural network (GNN) where nodes represent objects and edges represent the geometrical relationships between objects. We train our GNN model on the Visual Genome dataset, which contains over 100,000 natural images with object and relationship annotations. During the search, the trained GNN updates its prior knowledge about object relationships based on newly detected objects in the scene to make more accurate predictions. Our approach circumvents scalability issues present in traditional methods by implicitly encoding object relationships extracted from a large-scale real-world dataset in a flexible GNN model.

Toward the goal of fully autonomous assistant robots for everyday use, we further integrate our task planner with object navigation to execute human instructions. We build upon a hybrid task planner~\cite{SGL2022} that can generate task-completing action sequences based on human instructions. But the task planner assumes all required objects are located within the current field of view (FoV) of robot. Combining with the GNN-based object navigation, we enable the task planner to generate plans for searching missing objects that are outside of the current FoV. Our framework is illustrated in \cref{fig:teaser}. We assess the effectiveness of our proposed pipeline through simulations and real-world experiments. To benchmark the simulation results, we conduct object navigation in four room types, namely kitchen, bedroom, living room and bathroom, and four demanding tasks -- cut, cook, clean, and pick-and-place -- in AI2-THOR \cite{AI2THOR}, a realistic simulator of interactive household environments. Furthermore, we showcase that our pipeline can deliver reliable performance when applied to a real mobile manipulator Stretch RE2 robot \cite{RE2}.

\section{Related Works}

\textbf{Object Navigation} In previous works, object navigation is defined as the task of navigating to an object (specified by its category label) in an unexplored environment \cite{batra2020objectnav}. One popular approach is to use end-to-end reinforcement or imitation learning to build episodic memory and learn semantic priors implicitly \cite{mousavian2019visual,gupta2017cognitive}. However, these methods struggle at this task as they are ineffective at exploration and long-term planning. To improve the efficiency of exploration and searching, approaches like language guidance \cite{krantz2020beyond} and semantic map construction \cite{chaplot2020object} are proposed and used. Another parallel line of works have studied using partially observable Markov decision process (POMDP) to maintain and update a measure of uncertainty over the location of the target object, via its belief state \cite{8793494,zheng2022towards,holzherr2021efficient}. This formalization is useful because object search over long horizons is naturally a sequential, partially observed decision-making problem. But current POMDP-based approaches for object search often assume object independence to maintain and reason about belief states, which enables scalability but may compromise the accuracy of the search \cite{holzherr2021efficient}. Alternatively, these approaches may rely on accurate object correlation information, which may not always be readily available. In such cases, the success rate of the search process may significantly decrease \cite{zheng2022towards}. In contrast, our approach does not require prior knowledge of object correlation and addresses the scalability issue by implicitly encoding relationships inside the GNN model. This is achieved by training the model with a large-scale object relationship dataset. Therefore, our approach does not suffer from performance decrease due to inaccurate object correlation information.

\textbf{Task Planning from Human Instruction} Based on manually defined symbols, early symbolic planners exploited the syntactic structure of language to understand human instructions and statically generated a sequence of actions \cite{symbolic2}\cite{symbolic3}\cite{symbolic4}. However, this type of approach fails to interpret the diverse human instructions nor captures semantic meaning in incomplete sentences. Learning-based methods can automatically learn linguistic features via deep neural networks, eliminating the need for engineered symbolic structures to process natural language \cite{2018bert}\cite{brown2020language}. However, it remains challenging to plan a sequence of admissible actions through end-to-end learning approaches. Recent studies show that well-trained large language models (LLM) already contain information necessary to work as task planner when prompted with a single fixed example of a task description and its associated sequence of actions \cite{huang2022language}\cite{ahn2022can}\cite{huang2022inner}. However, due to the open-ended nature of the LLM, it is difficult to guarantee the success of generated plan. To ensure a task-completing plan, we leverage the strengths of symbolic and learning based approaches which adopts neural networks for goal learning and symbolic approaches for task planning.

\section{Approach}
\begin{figure*}
     \centering
     \includegraphics[width=0.9\textwidth,height=5cm]{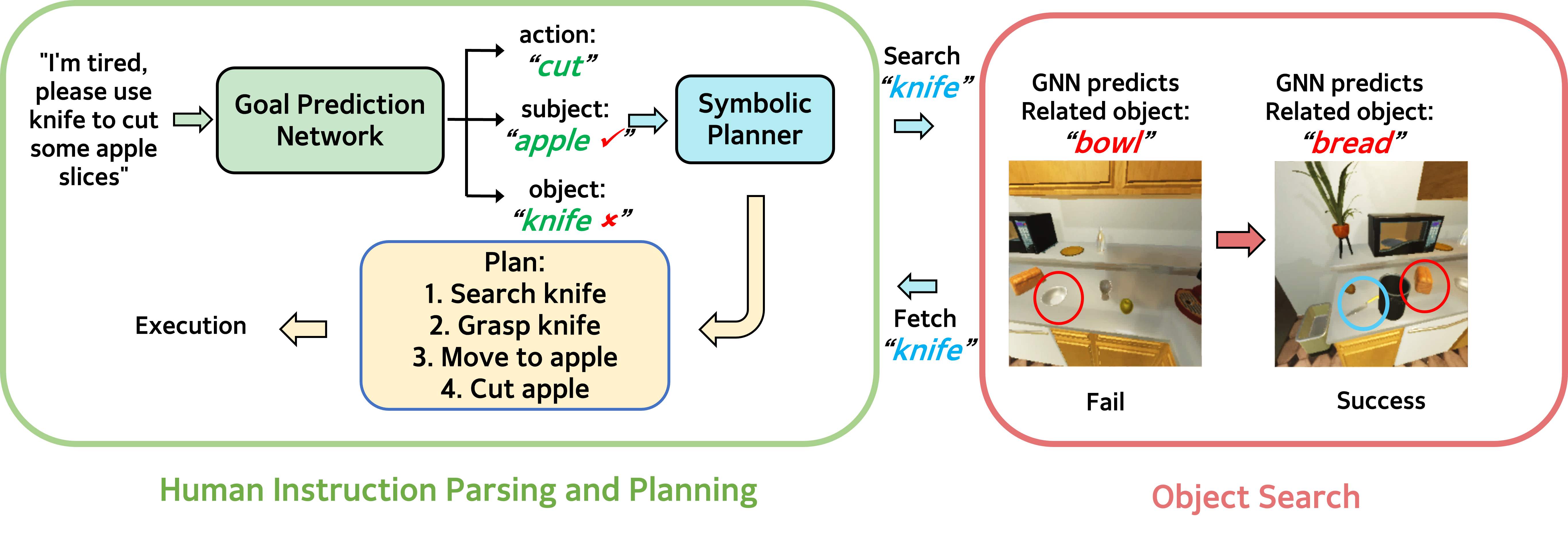}
     \caption{\textbf{Illustration of the proposed human instruction following pipeline}. The query from human will be parsed into a set of action, subject and object. To search for the missing \texttt{knife}, GNN predicts possible areas, such as \texttt{bowl} and \texttt{bread}, based on object co-occurrence relationships. Once the \texttt{knife} is retrieved, the robot moves back to \texttt{apple} and complete the action sequence generated via symbolic planner.}
     \label{fig:pipeline}
\end{figure*}

In this section, we will first introduce our solution for object navigation, followed by a description of the whole instruction following pipeline for assistant robots (\cref{fig:pipeline}). This pipeline includes task planning from human instructions as well as the object navigation algorithm.

\subsection{Object Navigation using Relationship Priors}
\label{ssec:nav}
\begin{figure}
     \centering
     \includegraphics[width=1\columnwidth]{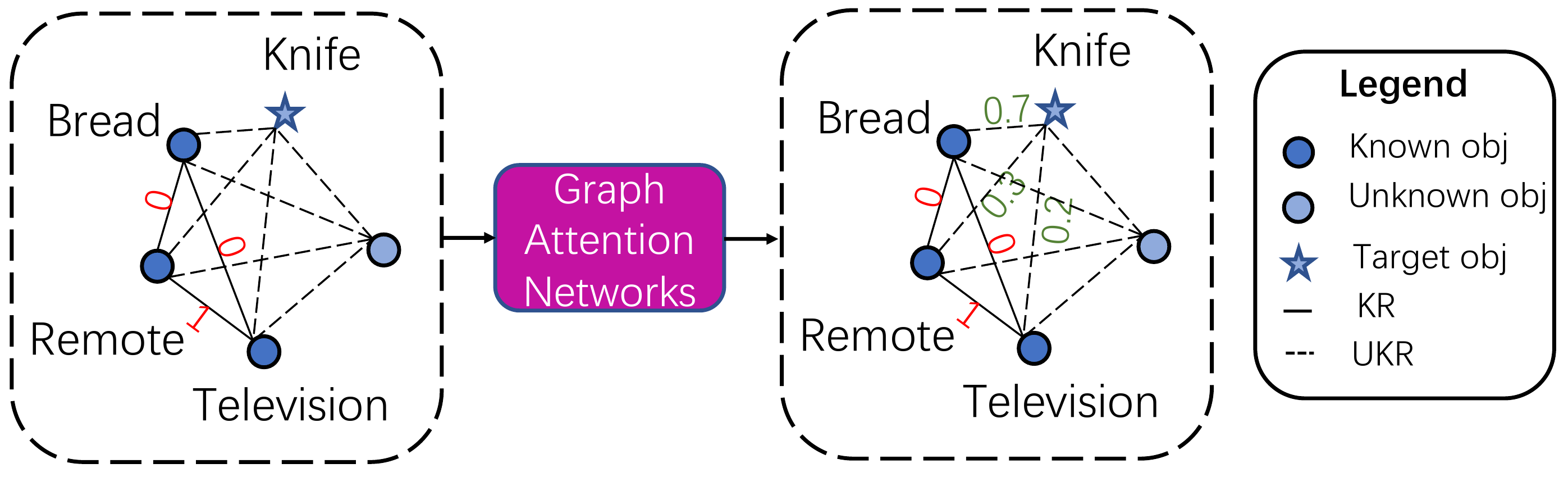}
     \caption{Illustration of correlation prediction. Given the current graph $G$, GNN predicts the probabilities of edge existence between target object \texttt{knife} and other known objects. KR means known relationship and UKR means unknown relationship.}
     \label{fig:GNN}
\end{figure}

\textbf{Problem Formulation} The problem is for a robot to search for a object with a specific category within an environment containing stationary and movable objects. The robot has access to a discrete 2D grid map of the environment that contains information about the positions of the $n$ stationary objects (e.g., furniture and fixtures), denoted by $x_1,\ldots, x_n$. However, the positions of the target object, denoted by $\xtarget$, and the $m$ movable objects, denoted by $y_1,\ldots, y_m$, are initially unknown to the robot. The objective is to efficiently locate the target object with the minimum number of steps.

\textbf{Object Search as object relationship prediction.} When searching for $\xtarget$, we aim to exploit the knowledge of the environment, such as the locations of stationary objects and detected moveable objects during navigation, and leverage the co-occurrence between object pairs to facilitate the search. Object searching is then formulated as iteratively predicting related objects whose positions are known to the robot until we find out $\xtarget$. To model, learn, and predict object relationships, we use the Graph Neural Network (GNN) structure, which is a powerful tool for processing graph data and modeling correlations \cite{xu2018powerful}. We denote our graph by $G = (V, E)$, where $V$ and $E$ denote the nodes and the edges between nodes, respectively. Specifically, each node $v \in V$ denotes an object category, and each edge $e \in E$ denotes a relationship between a pair of object categories. The input to each node $v$ is a feature vector $x_v$, which is the embedding of semantic object category. The goal is to learn to predict whether an edge exists between any two nodes.

\textbf{Learning object relationships from real-world data.} Visual Genome dataset \cite{Visual_Genome} enables us to learn real-world object correlations. This dataset contains over 100,000 natural images with object and relationship annotations. We can count the occurrence of spatial relationships between object pairs in the dataset, such as a bowl sitting in a sink, and connect two nodes if the occurrence frequency of any relationship is more than three. To compute node embeddings $h_v$, we use the Graph Attention Networks (GAT) architecture \cite{GAT}. GAT allows us to implicitly specify weights to neighbor nodes. We compute the node embeddings using $h_v = GAT(x_v, G)$. After we get the node embeddings, another prediction header would compute the probability of link existence, $\hat{z}_{uv} = Predictor(h_u, h_v)$, see \cref{fig:GNN}. 

\textbf{Search-time object relationship update.} During testing, the robot only has the location of $n$ stationary objects $x_1,\ldots, x_n$ at the start. The graph $G$ contains the relationships between these $n$ objects. The GNN model predicts the probability of link existence $E(\xtarget,x_1) \ldots, E(\xtarget,x_n)$ between the target object $\xtarget$ and each of the stationary objects $x_1,\ldots, x_n$. The model selects the object with the highest probability and searches its surrounding area. During navigation, the moveable objects in $y_1,\ldots, y_m$ will be detected and their locations will be estimated (assume rgbd image is available), thus new relationship would be computed and added into the graph $G$. This process allows the GNN to update its estimation of the object relationships for the environment and improve its predictions for the related object of $\xtarget$. We repeat the above process until the robot finds the target object $\xtarget$ or reaches the step limits. 


\subsection{Object Search to Follow Human Instruction} 
\label{ssec:inst}
In this subsection, we present the complete pipeline for following human instructions, which includes task planning and object navigation, as illustrated in the \cref{fig:pipeline}. A task-completing plan necessitates the successful execution of human instructions. We build upon the hybrid task planner that uses vision-and-language network to learn the goal symbol from the instruction, and then sending it to a symbolic planner to generate a task-completing plan \cite{SGL2022}.  However, this approach assumed that all required objects were always within the current FoV of the robot.

Our approach removes the reliance on images for task planning and eliminates the assumption that all required objects are within the FoV of the robot. To provide more detail, our approach begins with a natural language sentence $L$ composed of $K$ tokens, and outputs a sequence of primitive actions (e.g., search object, grasp object, and cut object) that will reach the desired goal state upon execution, see green box in \cref{fig:pipeline}. In detail, the goal state is parsed by a goal prediction network which predict the action $a$, subject $s$, and object $o$. Then, the task planning problem is addressed by transferring from an initial state to the desired goal state, using the Planning Domain Definition Language (PDDL) \cite{aeronautiques1998pddl}, a commonly used symbolic planning language. The PDDL planner produces a task-completing action sequence if the domain and problem are well-defined and specified, which includes a list of pre-defined objects and their corresponding predicates (e.g., "dirty" or "graspable"), as well as primitive actions and their corresponding effects. If either the required subject $s$ or object $o$ is missing, an additional searching step is added to the plan to locate the missing object using the navigation algorithm discussed above,  shown in the red box in \cref{fig:pipeline}. This allows our pipeline to extend the capabilities of home-assistant robots by enabling them to search for objects if they are missing, making robots more flexible and adaptable in real-world scenarios.


\section{Experimental Results}
Our experiments aim to show that (1) Object co-occurrences relationship can be captured using GNN model and used for efficient object navigation; (2) The integration of object navigation with a task planner enables home assistant robots to successfully complete everyday household tasks based on human instructions. We demonstrate the above two key points through the experiments in AI2-THOR simulator and the physical Stretch robot.

\subsection{AI2-THOR Simulation Results}
The AI2-THOR \cite{AI2THOR} simulator provides accurate modeling of 120 interactive scenes, evenly split between kitchens, living rooms, bedrooms, and bathrooms. It includes hundreds of unique objects that have visual state changes such as open/close, on/off, and clean/dirty. The robot can take primitive move actions from the set: $\{\texttt{MoveAhead}, \texttt{RotateLeft},\texttt{RotateRight}\}$. \texttt{MoveAhead} moves the robot forward by 0.25m. \texttt{RotateLeft}, \texttt{RotateRight} rotate the robot in place by 45$^\circ$. Moreover, robot can change the positions and states of objects through action primitives:  $\{\texttt{Pick}, \texttt{Put}, \texttt{Move}, \texttt{Cook}, \texttt{Slice},$ $\texttt{Toggle}$$\}$. 

\textbf{Zero-shot Object Link Prediction}: As described in sec.~\ref{ssec:nav}, we train our GNN model using the Visual Genome dataset. Here we evaluate the link prediction performance of the trained GNN model using both the Visual Genome test split and the AI2-Thor environment. We consider 94 unique object categories in AI2-Thor to construct the ground truth object relationship graph. A pair of objects are considered connected if they are in close proximity based on Euclidean distance or have receptacle relationship. The receptacle relationship indicates that one of the objects is a receptacle, and the other object is either inside or on top of it. An example of this relationship would be an apple located on a desk. The edge label is created using the following rule:

\begin{equation}
\begin{aligned}
  E(obj_i,obj_j)=&\begin{cases}
    1 & \norm{obj_i - obj_j} < d_{\text{thre}} \\
    1 & $is receptacle relation$ $($obj_i, obj_j$)$\\
    0 & $else$ \\
    \end{cases}\\
\end{aligned}
\label{eq:relation}
\end{equation}

where $obj_i$ and $obj_j$ are the locations of an object pair, and $d_{\text{thre}}$ is a predefined threshold (1.0m).

The link prediction performance is shown in tab.~\ref{tab:link_accuracy}. Our model achieved high accuracy in the Visual Genome test split and is able to transfer its knowledge to AI2-THOR environment and achieve a zero-shot prediction accuracy of $67\%$ (without the search-time update described in sec.~\ref{ssec:nav}). We hypothesize that this is because the AI2-THOR scenes closely resemble real-world environments, and the Visual Genome dataset provides a vast amount of data to aid in learning object relationships in such environments. Overall, these results support our hypothesis that object co-occurrences can be effectively captured with GNN models trained on real-world scene dataset.

\begin{table}[htbp]
\centering
\caption {Link Prediction Accuracy in Visual Genome dataset and AI2-THOR Scenes.}
\label{tab:link_accuracy}
\begin{tabular}{ccc}
\hline
  & Visual Genome test split & AI2-THOR scenes\\ 
\midrule
{Link Pred. Accuracy} & 86.66\% & 67.36\% \\
\hline
\end{tabular}
\end{table}

\begin{figure}[htbp]
\vspace{-5pt}
     \centering
     \begin{subfigure}{0.15\textwidth}
         \centering
         \includegraphics[width=\textwidth]{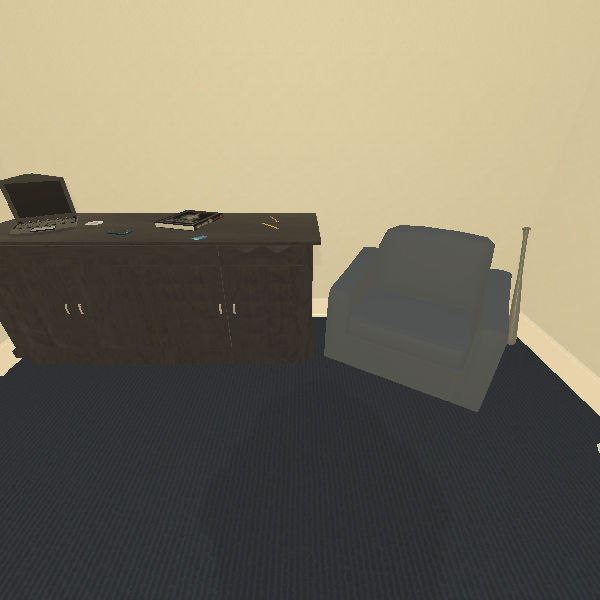}
     \end{subfigure}
     \begin{subfigure}{0.15\textwidth}
         \centering
         \includegraphics[width=\textwidth]{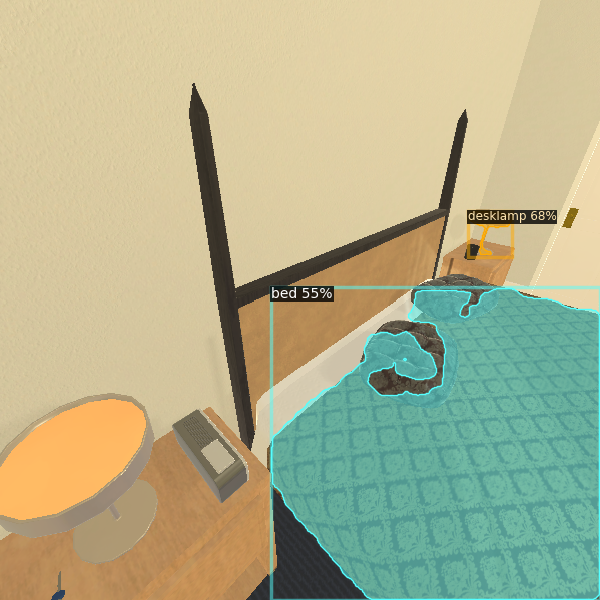}
     \end{subfigure}
     \begin{subfigure}{0.15\textwidth}
         \centering
         \includegraphics[width=\textwidth]{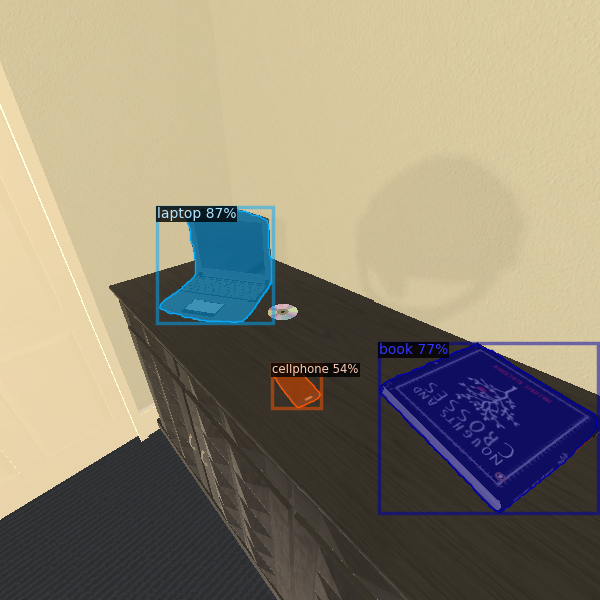}
     \end{subfigure}

     \hfill  
     \vspace{-4pt}
     
     \begin{subfigure}{0.15\textwidth}
         \centering
         \includegraphics[width=\textwidth]{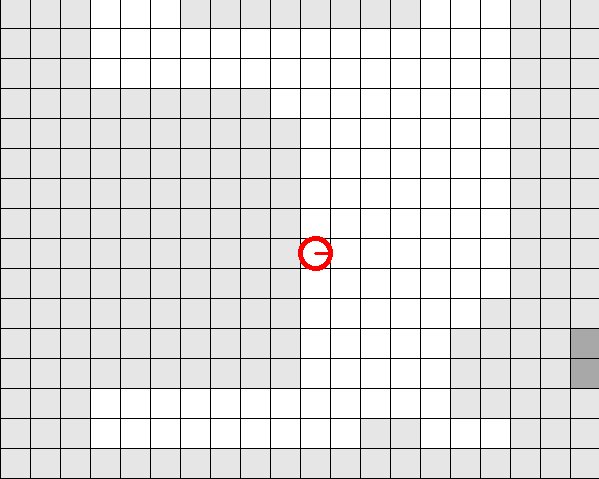}
     \end{subfigure}
     \begin{subfigure}{0.15\textwidth}
         \centering
         \includegraphics[width=\textwidth]{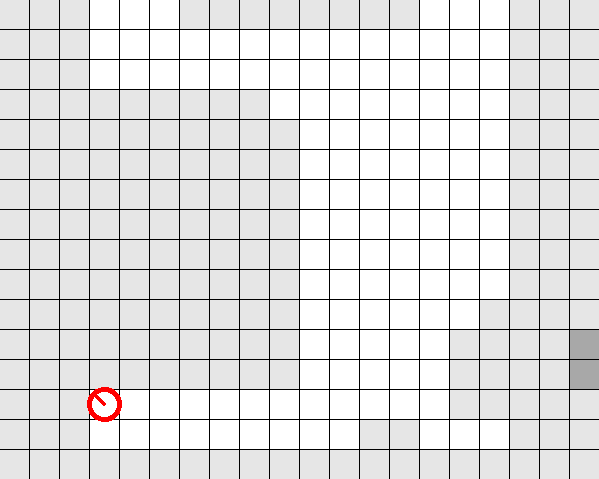}
     \end{subfigure}
     \begin{subfigure}{0.15\textwidth}
         \centering
         \includegraphics[width=\textwidth]{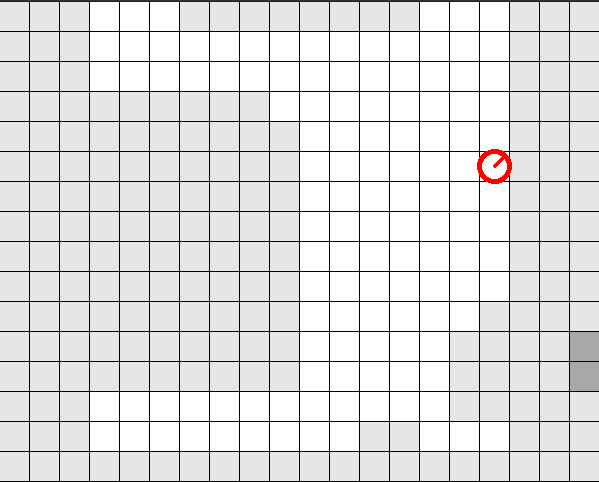}
     \end{subfigure}
     \caption{\textbf{Example sequence of searching for \texttt{book} in a living room scene.} Top: first-person view with object detection. Bottom: grid map of the environment. GNN first predicts \texttt{bed} as related but failed to find \texttt{book}, then it predicts \texttt{dresser table} and successfully finds the \texttt{book} on it.}
     \label{fig:search plot}
\end{figure}

\textbf{Object Navigation}: 
We compared our approach to Zheng et al.'s methods \cite{zheng2022towards}, COS-POMDP, Target-POMDP, and Greedy-NBV. All of these search approaches use object correlation, either based on ground-truth or estimated relationship. They maintain and update a probability measurement of the target object over its belief state for searching. We ensure a fair comparison by allowing all algorithms to take the same number of steps for searching and the same set of target object classes in test scenes. We also use the same YOLOv5 object detector \cite{glenn_jocher_2022_7347926} as the perception module for all methods. The robot's success condition is to detect the target object and be within 1.0m Euclidean distance from it by the end of an episode. As shown in~\cref{tab:navigation comparison}, our approach outperforms these baselines in both searching success rate and efficiency. The example process of object searching in AI2-THOR is shown in \cref{fig:search plot}.

It is important to note that the baseline models rely on accurate correlational information. In particular, these models label object pairs as "close" if their distance is smaller than 2.0m. When using estimated object correlation instead of the ground truth information, the performance of COS-POMDP (indicated by the ``est'' variant) declines significantly. This finding highlights the importance of leveraging large-scale knowledge source when object correlations are difficult to estimate during the search process. And our approach provides the complement for these methods and makes it suitable for a wide range of unseen environments without object correlations information. 


\begin{table*}[h!]
\centering
\caption {\textbf{AI2-THOR Object Navigation Quantitative Performance.} The task success rate (SR) and navigation efficiency which is the success weighted by path length (SPL) of our approach and a variety of prior algorithms on AI2-THOR.}
\label{tab:navigation comparison}
\begin{tabular}{@{}ccccccccc@{}}
\toprule
  & \multicolumn{2}{c}{Kitchen} & \multicolumn{2}{c}{Bathroom} & \multicolumn{2}{c}{Bedroom} & \multicolumn{2}{c}{Living room}\\
  Model & SPL & SR & SPL & SR & SPL & SR & SPL & SR\\ \midrule
{Greedy-NBV} & 11.61\% & 31.03\%& 14.34\% & 34.48\%& 16.92\% & 26.67\%&7.13\% & 20.00\%\\\midrule
{Target-POMDP} & 13.80\% & 34.48\%& 19.88\% & 55.17\%& 19.79\% &26.67\%& 24.36\% &40.00\%\\\midrule
{COS-POMDP}& 20.45\%& 41.38\% & 30.64\% & 55.17\% & 24.76\% & 40.00\% & 24.99\% & 43.33\%\\\midrule
{COS-POMDP (est)}& 8.39\%& 20.69\% & 17.20\% & 41.38\% & 16.78\% & 30.00\% & 14.07\% & 26.67\%\\\midrule
{GNN-Nav} & \textbf{23.56\%} & \textbf{46.67\%}& \textbf{56.62\%} & \textbf{86.67\%} & \textbf{28.91\%} & \textbf{53.33\%} & \textbf{35.01\%} & \textbf{59.25\%}\\
\bottomrule
\end{tabular}
\end{table*}

\textbf{Human Instruction Following}: Our ultimate goal is to enable a home-assistant robot to follow and execute human instruction. To verify the effectiveness of our human instruction following pipeline discussed in sec.~\ref{ssec:inst}, we conducted tests on four daily activities: cutting, cooking, cleaning, and pick-and-place. To complete the tasks, we utilized the default action space defined in AI2-THOR, which includes slice, toggle, and clean. Each task involves the robot receiving explicit human instruction \footnote{Explicit human instruction contains the subject and object inside which would make the training easier.}. Target objects that are required to complete the task are initially not in the FoV of robot. While the location of the subject is assumed to be known to the robot. The list of target and subject classes and the task descriptions are provided below. 

\begin{itemize}
\item \textbf{Pick-and-place.} \emph{Targets} are \texttt{apple}, \texttt{bread}, \texttt{fork}, \texttt{lettuce}, \texttt{potato}, \texttt{tomato}; \emph{Subjects} are \texttt{plate}, \texttt{bowl}, \texttt{sink}; \emph{Task} is pick the \emph{target} and place it on \emph{subject}.
\item \textbf{Cook.} \emph{Targets} are  \texttt{bread}, \texttt{egg}, \texttt{lettuce}, \texttt{potato}, \texttt{tomato}; \emph{subjects} are \texttt{microwave}, \texttt{toaster}, \texttt{stoveburner}; \emph{Task} is cook the \emph{target} with \emph{subject}.
\item \textbf{Clean.} \emph{Targets} are \texttt{mug}, \texttt{spatula}, \texttt{cup}, \texttt{butterknife}, \texttt{pan}, \texttt{bowl}, \texttt{plate}, \texttt{pot}; \emph{subject} is \texttt{sink}; \emph{Task} is clean the \emph{target} inside the \emph{subject}.
\item \textbf{Cut.} \emph{Targets} are \texttt{bread}, \texttt{lettuce}, \texttt{potato}, \texttt{tomato}; \emph{subject} is \texttt{knife}; \emph{Task} is cut the \emph{target} with \emph{subject}.
\end{itemize}

We carry out evaluation in 30 distinct kitchen environments, and our criterion for successful completion of a human instruction task involved several factors. Specifically, the robot must correctly parse the instructions and plan the task-completing action sequence (task planning), search for the target object (object navigation), and accurately identify the relevant subjects for performing the necessary actions (perception and manipulation).

In detail, we use the Symbolic Goal Learning Dataset\footnote{https://smartech.gatech.edu/handle/1853/66305}, and select 8163 explicit human instructions to train and test the language model. We adopt the Multi Modal Framework (MMF) \cite{mmf} and only train the LSTM language encoder with human instructions and corresponding ground-truth goal states. Our goal prediction network achieves $100\%$ prediction accuracy in 1024 unseen explicit human instructions, and the symbolic PDDL planner can create an optimal task plan once the goal state is correctly identified. For navigation and execution, the success rate is summarized in \cref{tab:hif_result}. There are two failure modes in our tests. The first mode is when the object navigation algorithm is unable to locate the target object within the step limit. The second failure mode arises when the robot is unable to find the required subject, such as a knife, due to either a perception error or additional actions being required, such as opening a drawer to retrieve the subject.

\begin{table}[htbp]
\centering
\caption {\textbf{AI2-THOR Human Instruction Following Quantitative Performance.} Nav means the success rate of navigation part, Exe means the success rate of task execution.}
\label{tab:hif_result}
\begin{tabular}{ccc}
\hline
 Task & Navigation (Nav) & Execution (Exe)\\ 
\midrule
{Cook} & 47.5\% & 42.5\% \\
{Clean}& 62.5\% & 62.5\% \\
{Cut}& 55.0\% & 20.0\% \\
{Pick-and-Place}& 47.5\% &45.0\%\\
\hline
\end{tabular}
\end{table}

\subsection{Real Robots Results}
\begin{figure}[htbp]
     \centering
     \begin{subfigure}{0.23\textwidth}
         \centering
         \includegraphics[width=\textwidth]{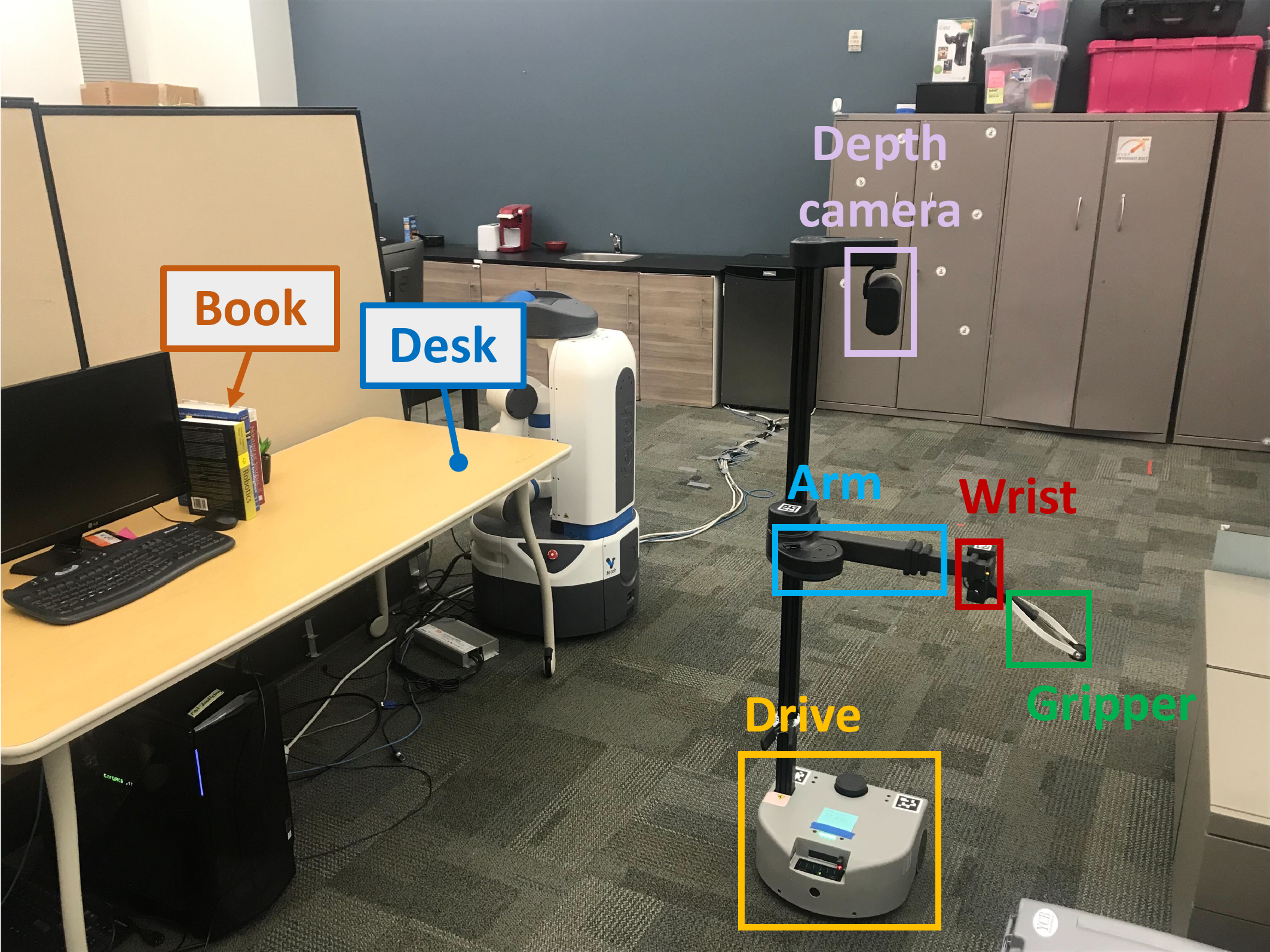}
     \end{subfigure}
     \begin{subfigure}{0.23\textwidth}
         \centering
         \includegraphics[width=\textwidth]{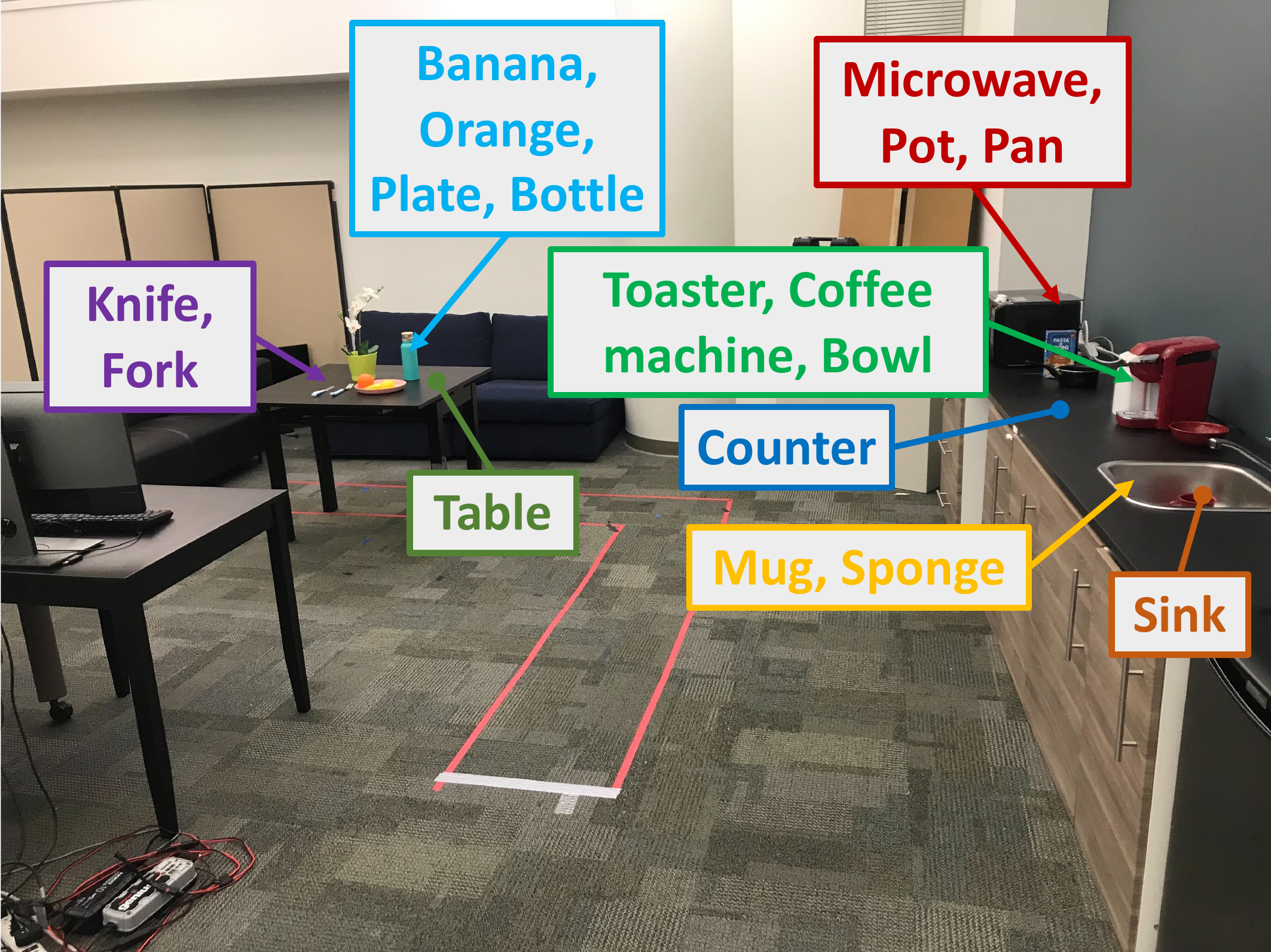}
     \end{subfigure}
     \caption{Real-world setup. Left: Robot. Right: Testing Scene.}
     \label{fig:robot and scene}
\end{figure}

In order to further demonstrate the effectiveness of our object navigation algorithm and human instruction following pipeline in real-world, we conduct experiments on a Stretch RE2 robot, which is specifically designed for navigation and manipulation in human environments. To create a realistic household scene, we place moveable objects in semantically related areas, as shown in \cref{fig:robot and scene}. We also provide the robot with the locations of stationary objects, including the table, desk, counter, sink, sofa, and microwave. Moreover, to prevent the perception module from detecting target objects that are far away and sidestepping the search process, we create obstructions such that the robot can only detect the object when being physically close to the object. We adopt the Detic open-vocabulary~\cite{Detic} object detector as the perception module and use the built-in depth camera to perform 3D pose estimation for the subsequent navigation and manipulation decisions. 

\begin{figure*}[htbp]
     \centering
     \begin{subfigure}{0.18\textwidth}
         \centering
         \includegraphics[width=\textwidth]{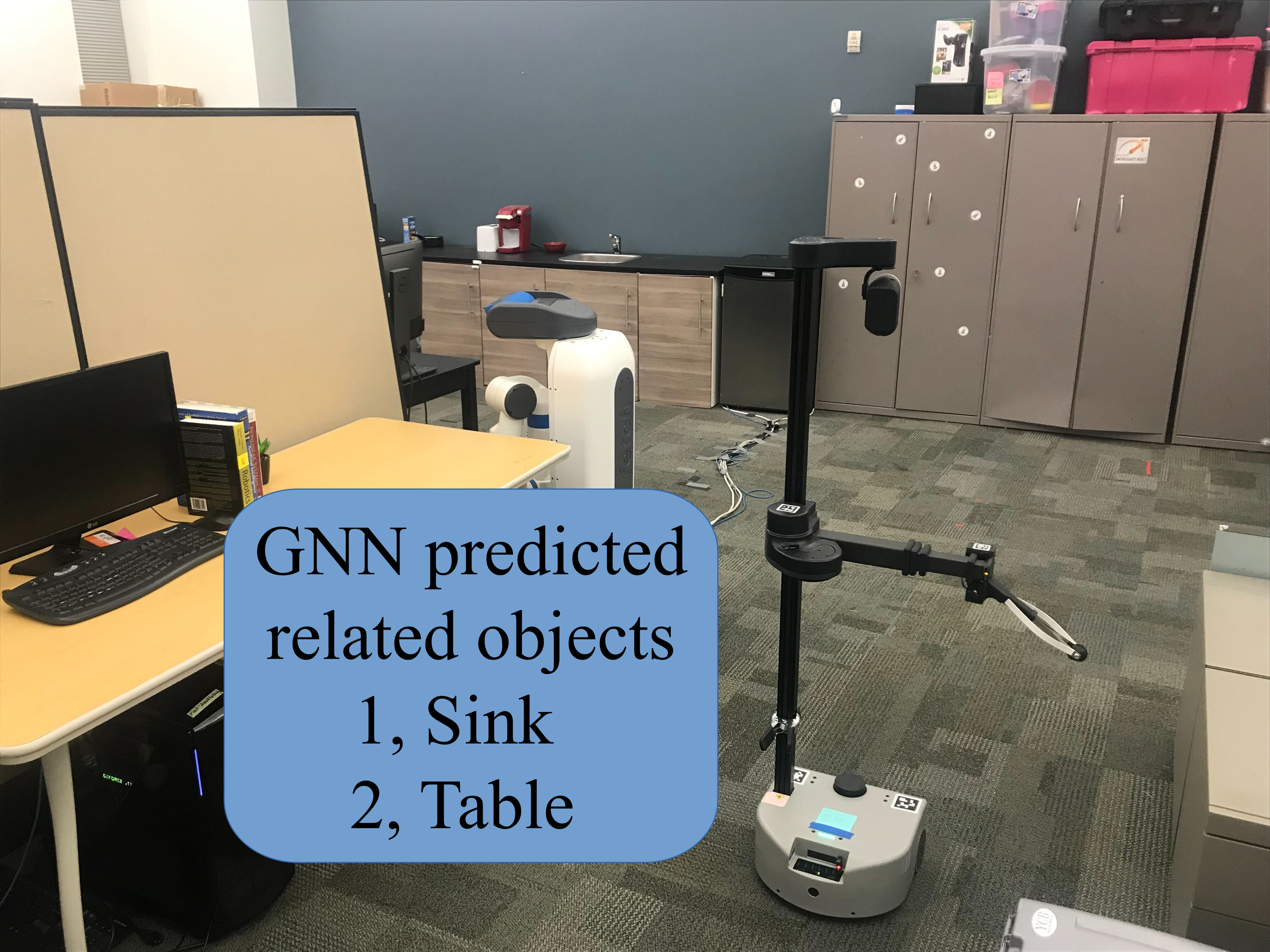}
         \caption{Predict related object}
     \end{subfigure}
     \begin{subfigure}{0.18\textwidth}
         \centering
         \includegraphics[width=\textwidth]{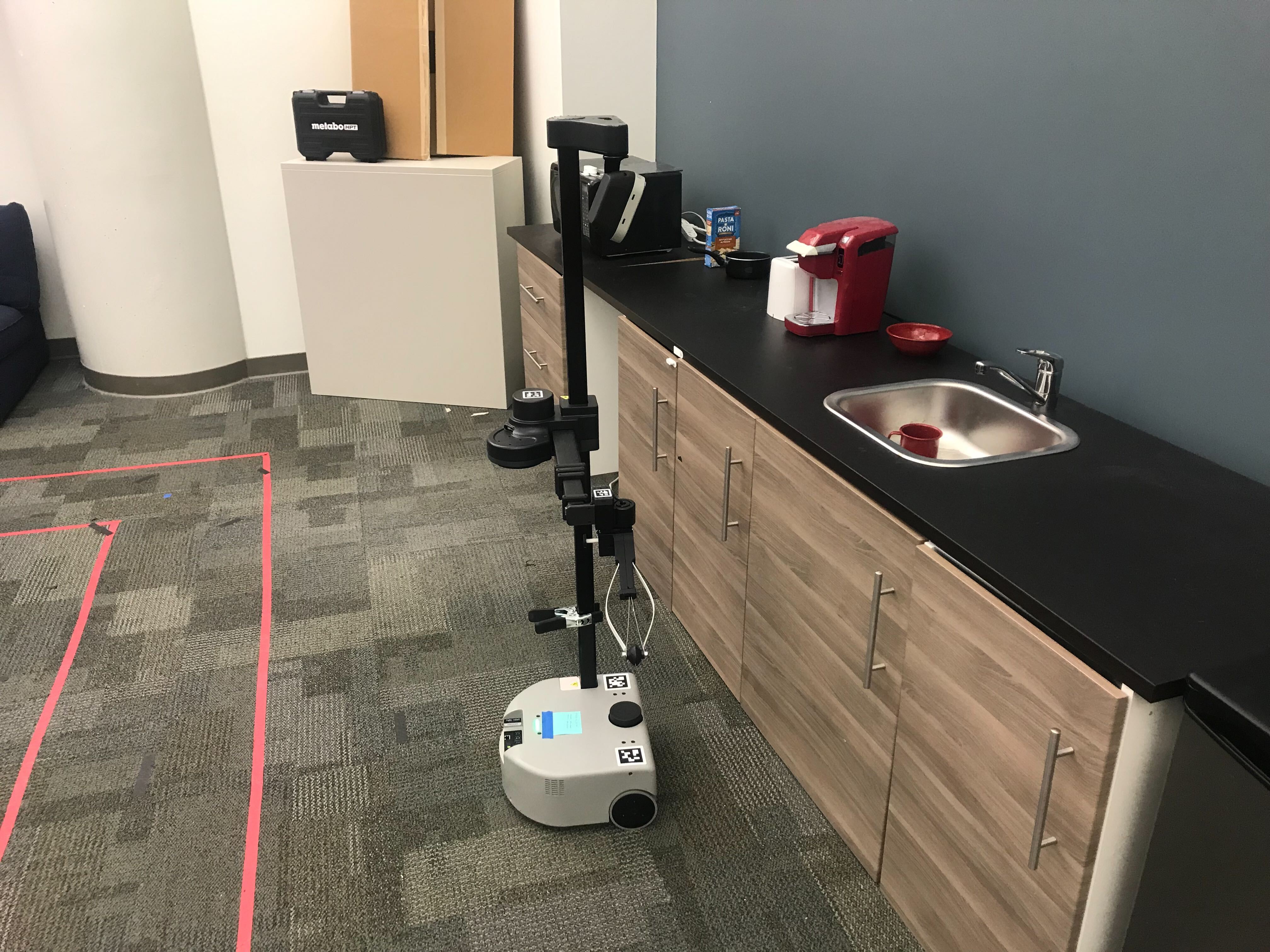}
         \caption{Search sink}
     \end{subfigure}
     \begin{subfigure}{0.18\textwidth}
         \centering
         \includegraphics[width=\textwidth]{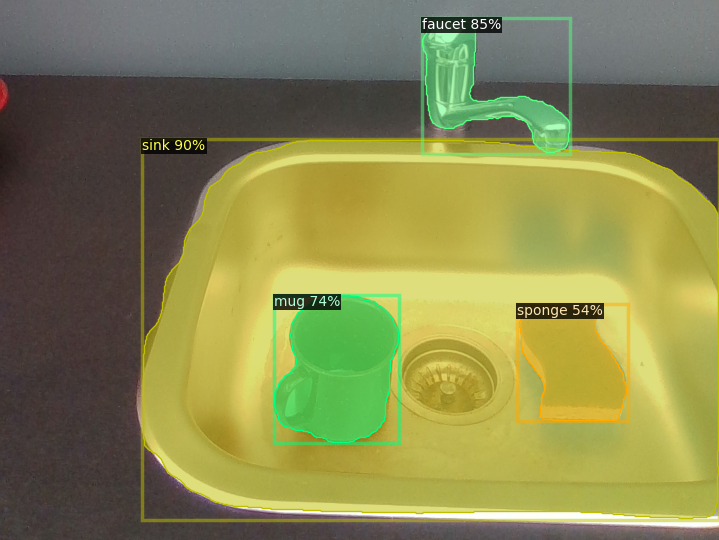}
         \caption{Sink detection}
     \end{subfigure}
     \begin{subfigure}{0.18\textwidth}
         \centering
         \includegraphics[width=\textwidth]{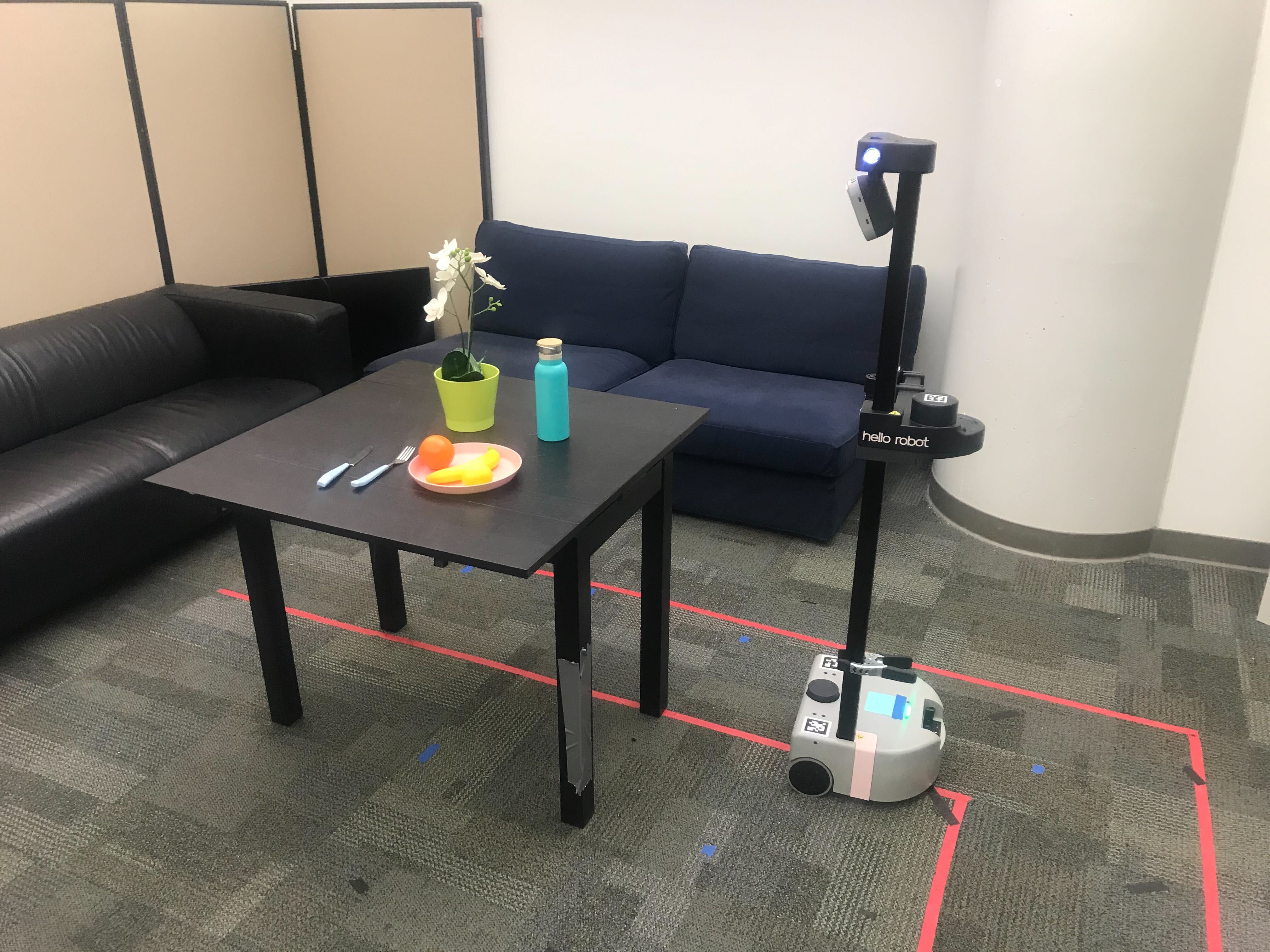}
         \caption{Search table}
     \end{subfigure}
     \begin{subfigure}{0.18\textwidth}
         \centering
         \includegraphics[width=\textwidth]{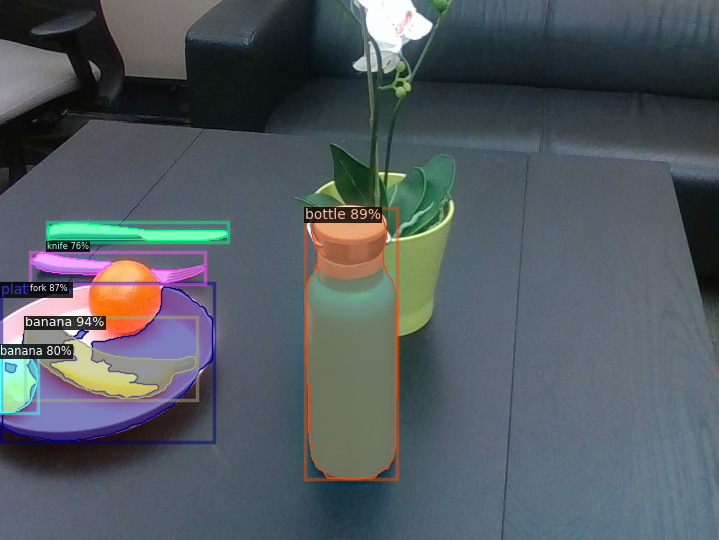}
         \caption{Table detection}
     \end{subfigure}
     \caption{Real world object navigation test (search for bottle).}
     \label{fig:success real world object search}
\end{figure*}

\begin{figure*}[htbp]
     \centering
     \begin{subfigure}{0.18\textwidth}
         \centering
         \includegraphics[width=\textwidth]{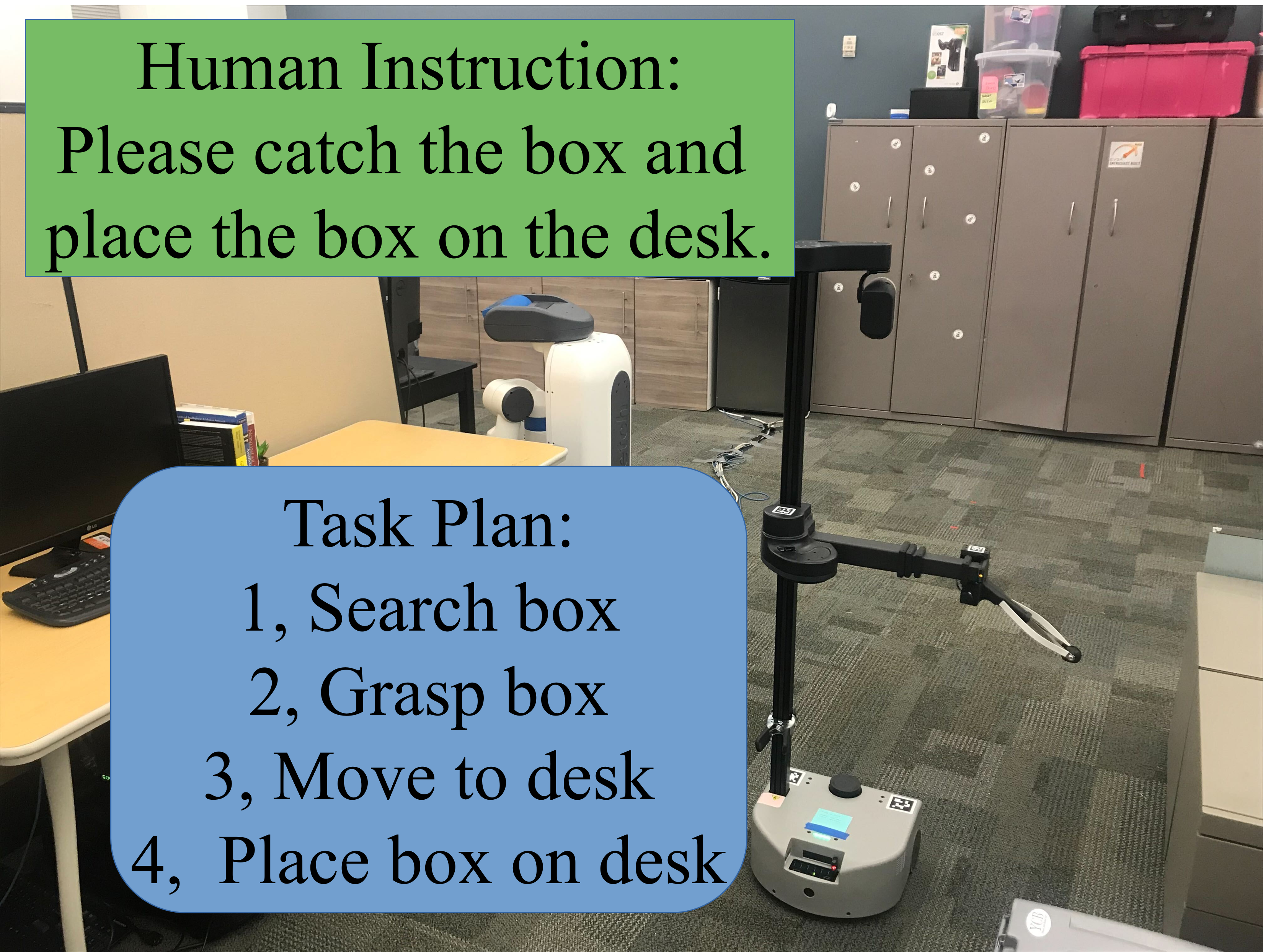}
         \caption{Task planning}
     \end{subfigure}
     \begin{subfigure}{0.18\textwidth}
         \centering
         \includegraphics[width=\textwidth]{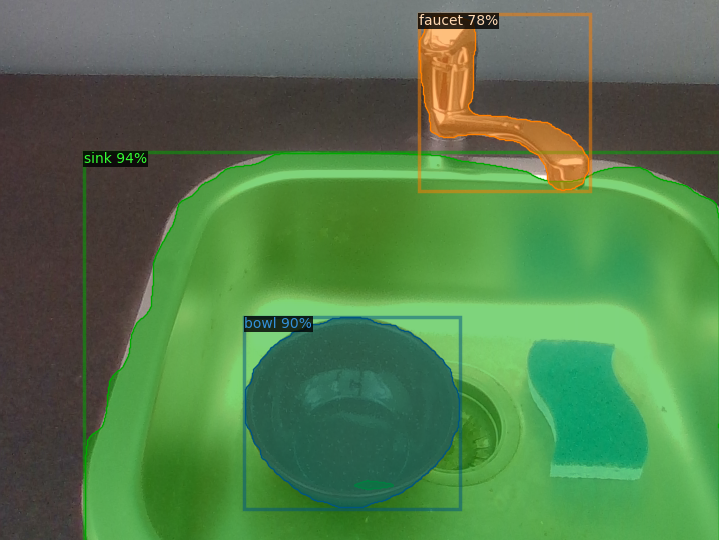}
         \caption{Search sink}
     \end{subfigure}
     \begin{subfigure}{0.18\textwidth}
         \centering
         \includegraphics[width=\textwidth]{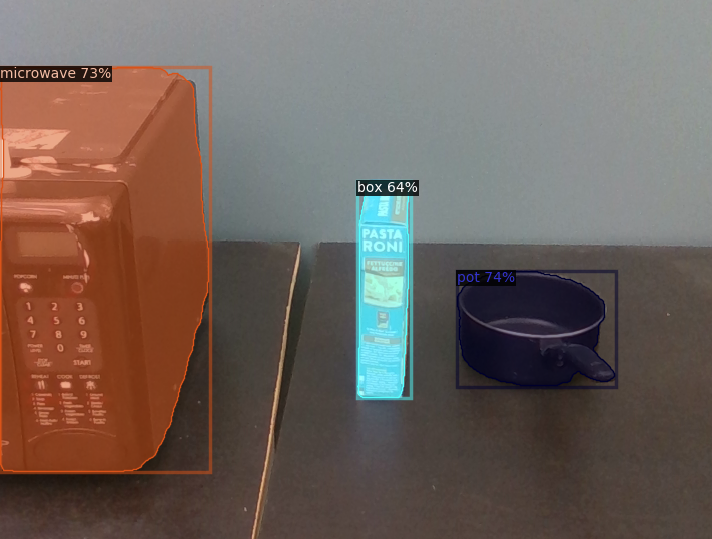}
         \caption{Search counter}
     \end{subfigure}
     \begin{subfigure}{0.18\textwidth}
         \centering
         \includegraphics[width=\textwidth]{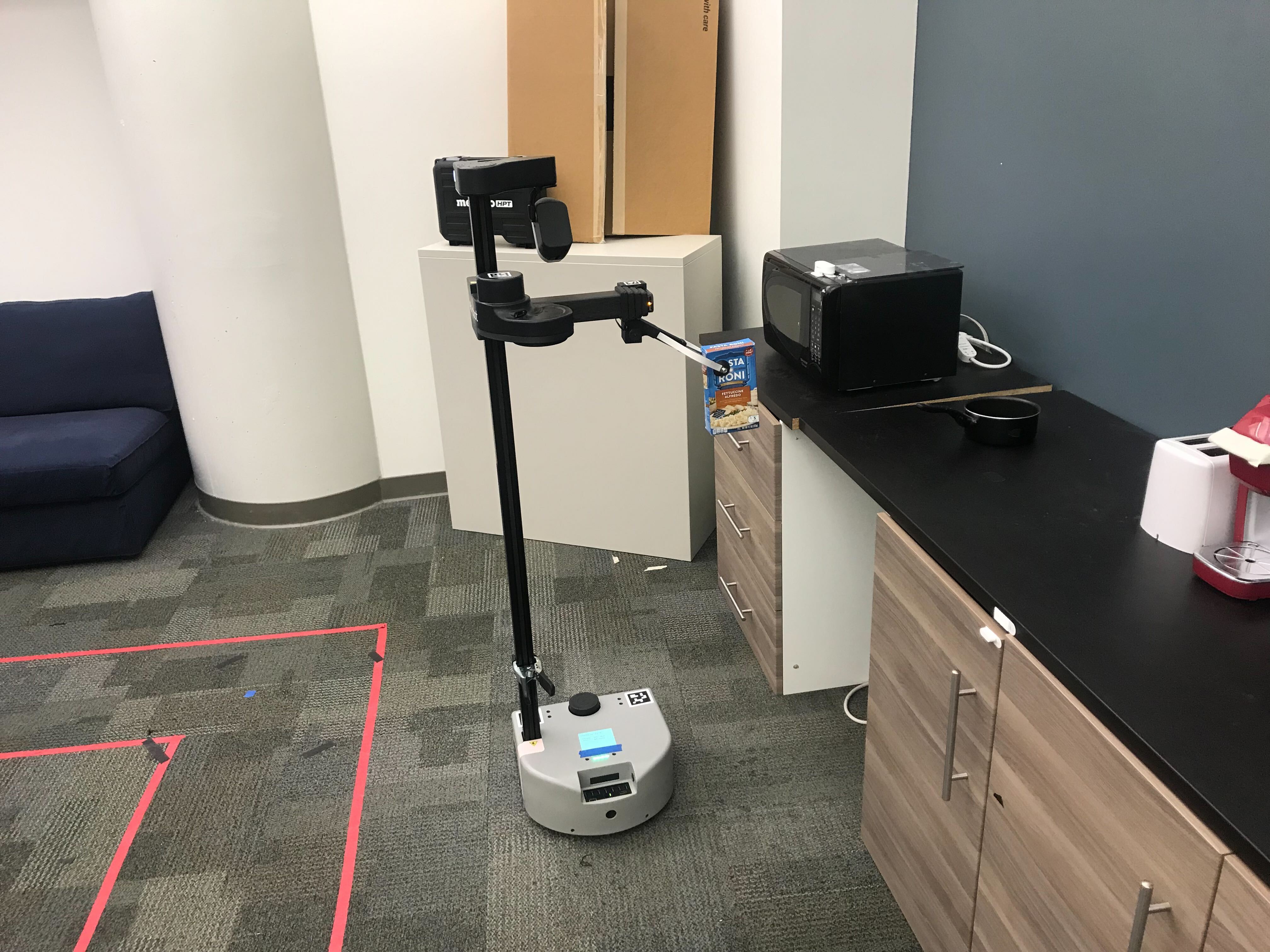}
         \caption{Grasp box}
     \end{subfigure}
     \begin{subfigure}{0.18\textwidth}
         \centering
         \includegraphics[width=\textwidth]{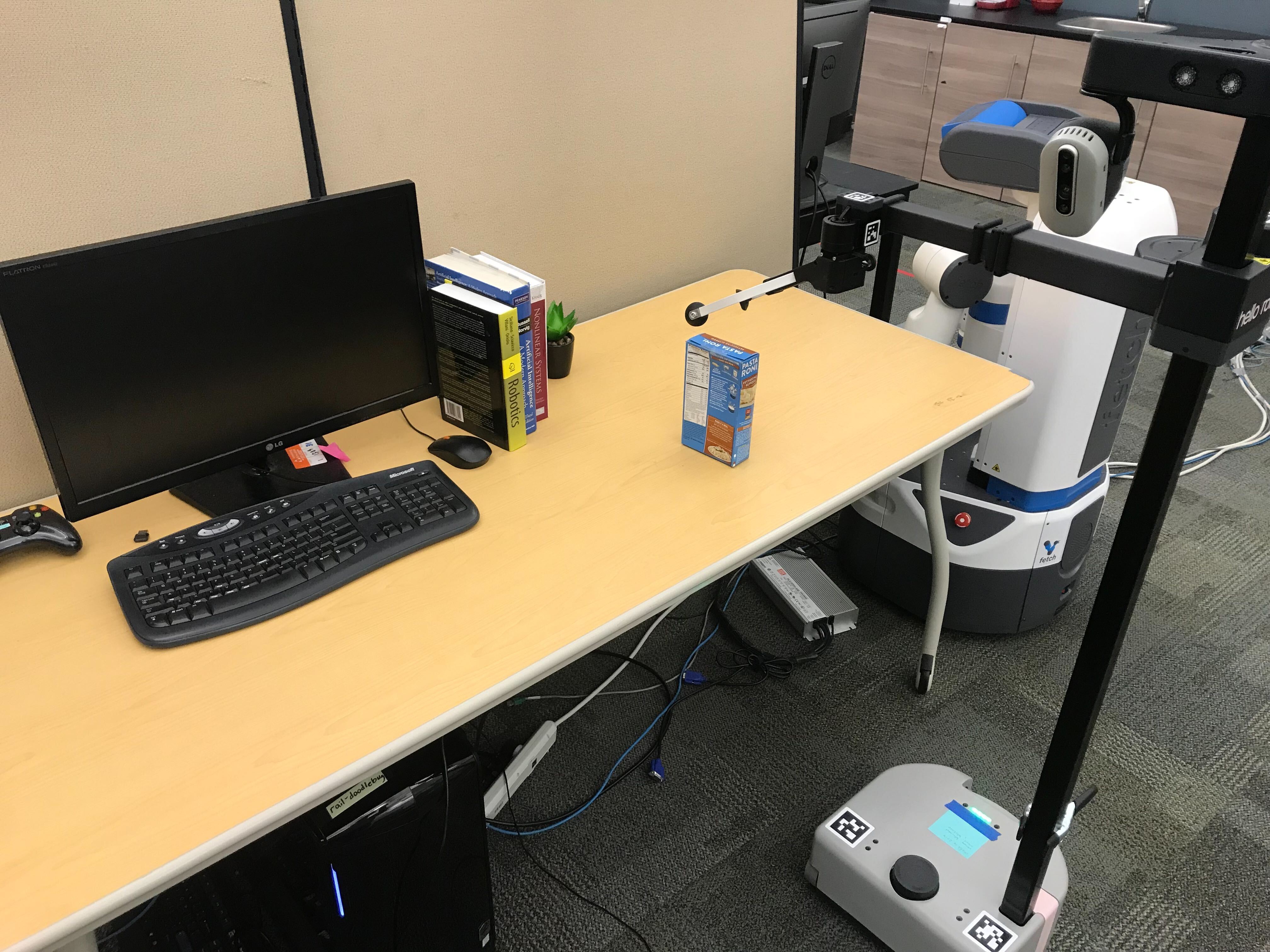}
         \caption{Place box}
     \end{subfigure}
     \caption{Real world human instruction following (pick\&place test).}
     \label{fig:hif_pp_test}
\end{figure*}

\begin{figure*}[htbp]
     \centering
     \begin{subfigure}{0.18\textwidth}
         \centering
         \includegraphics[width=\textwidth]{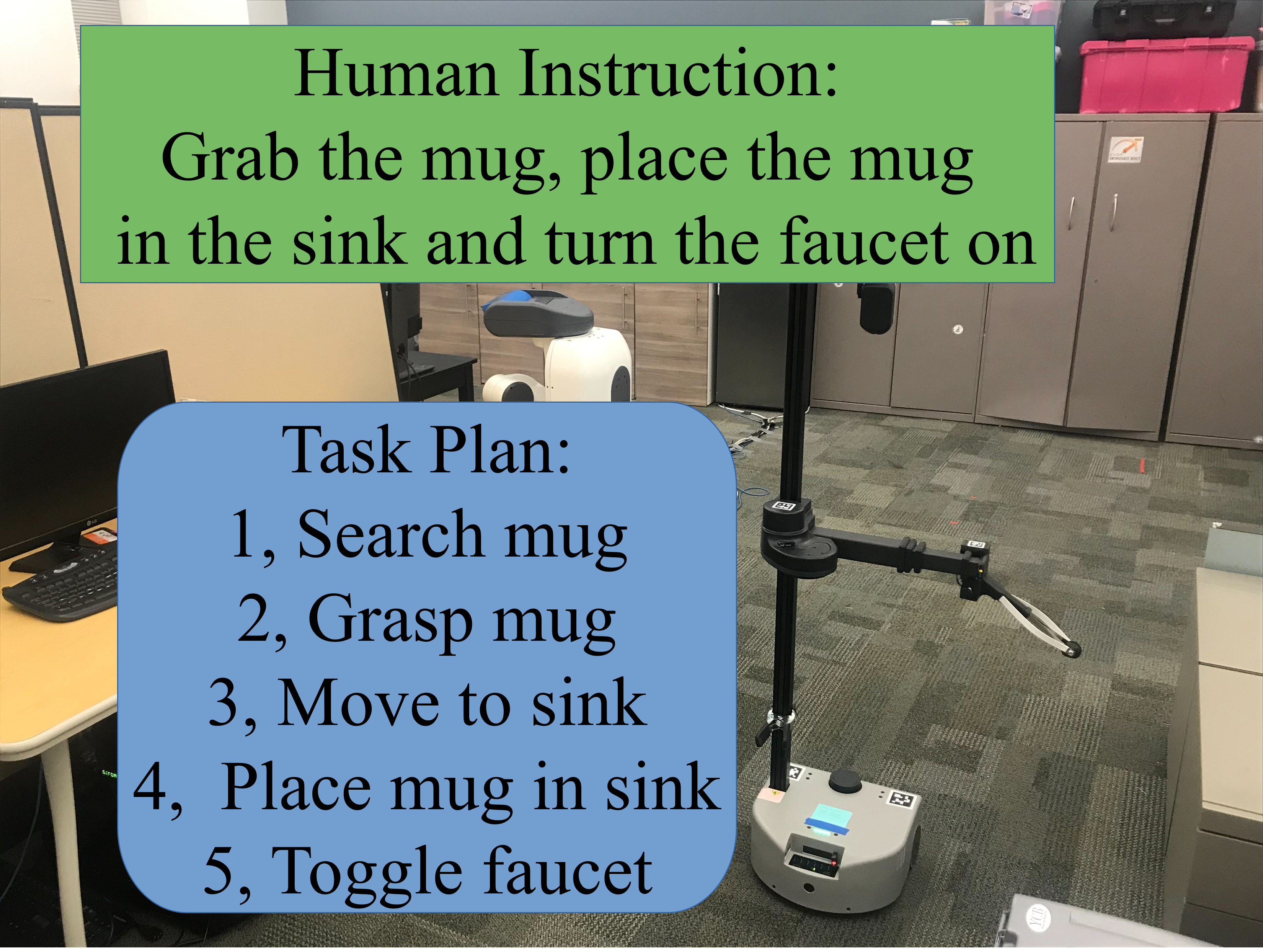}
         \caption{Task planning}
     \end{subfigure}
     \begin{subfigure}{0.18\textwidth}
         \centering
         \includegraphics[width=\textwidth]{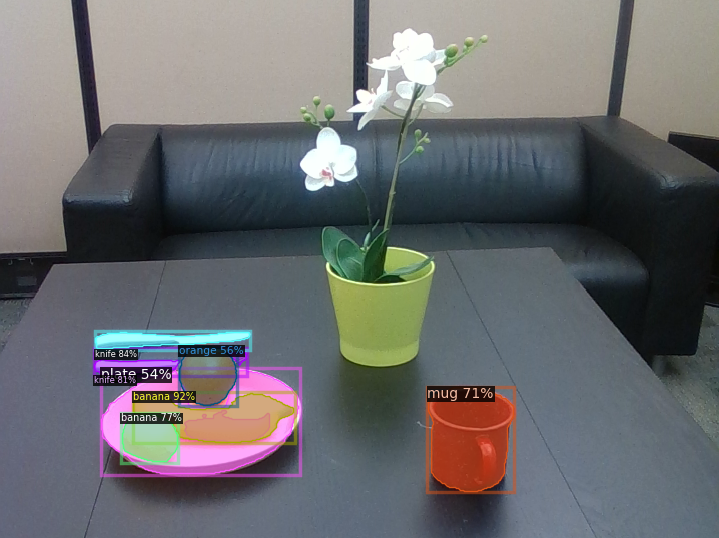}
         \caption{Search table}
     \end{subfigure}
     \begin{subfigure}{0.18\textwidth}
         \centering
         \includegraphics[width=\textwidth]{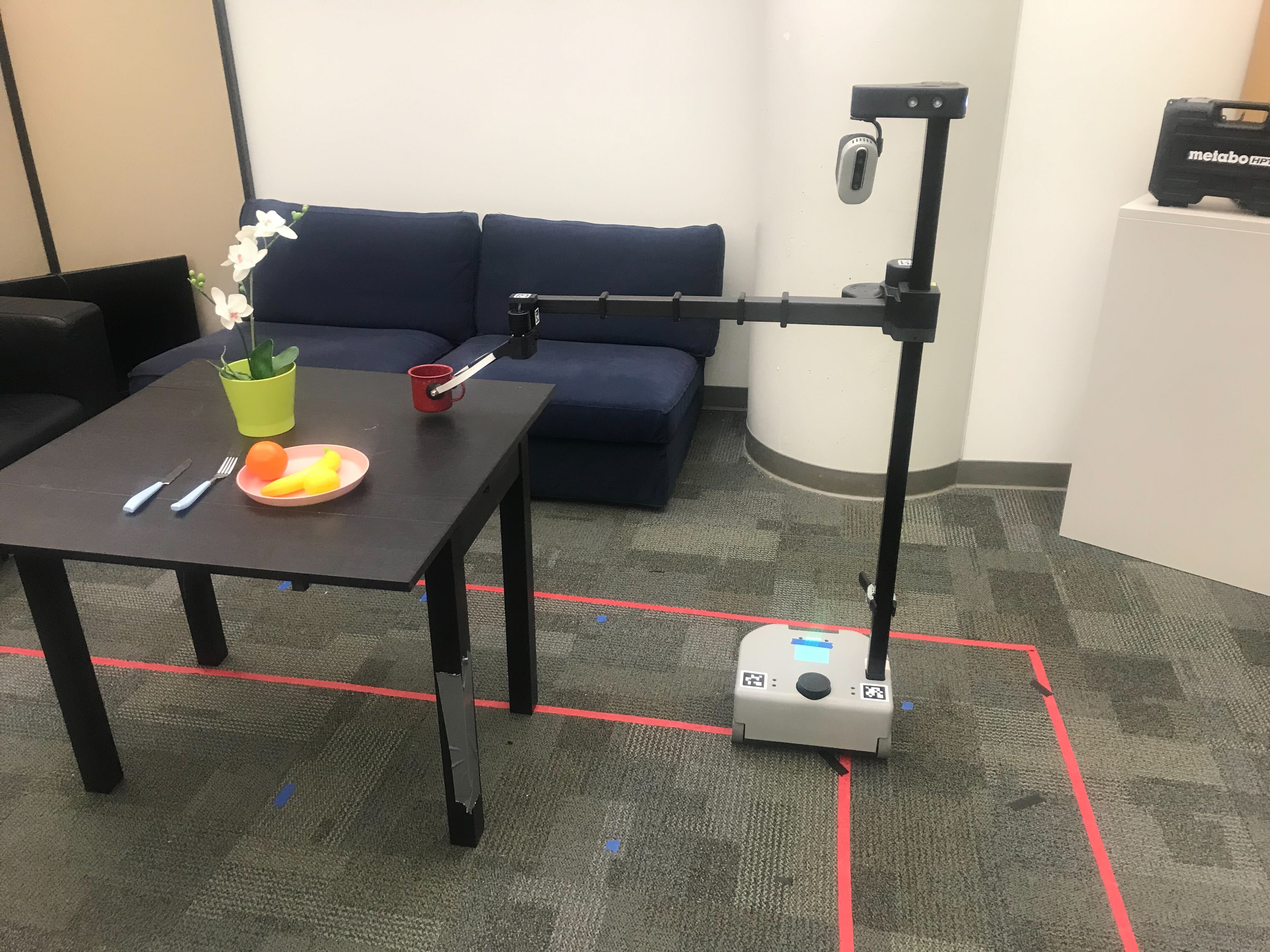}
         \caption{Grasp mug}
     \end{subfigure}
     \begin{subfigure}{0.18\textwidth}
         \centering
         \includegraphics[width=\textwidth]{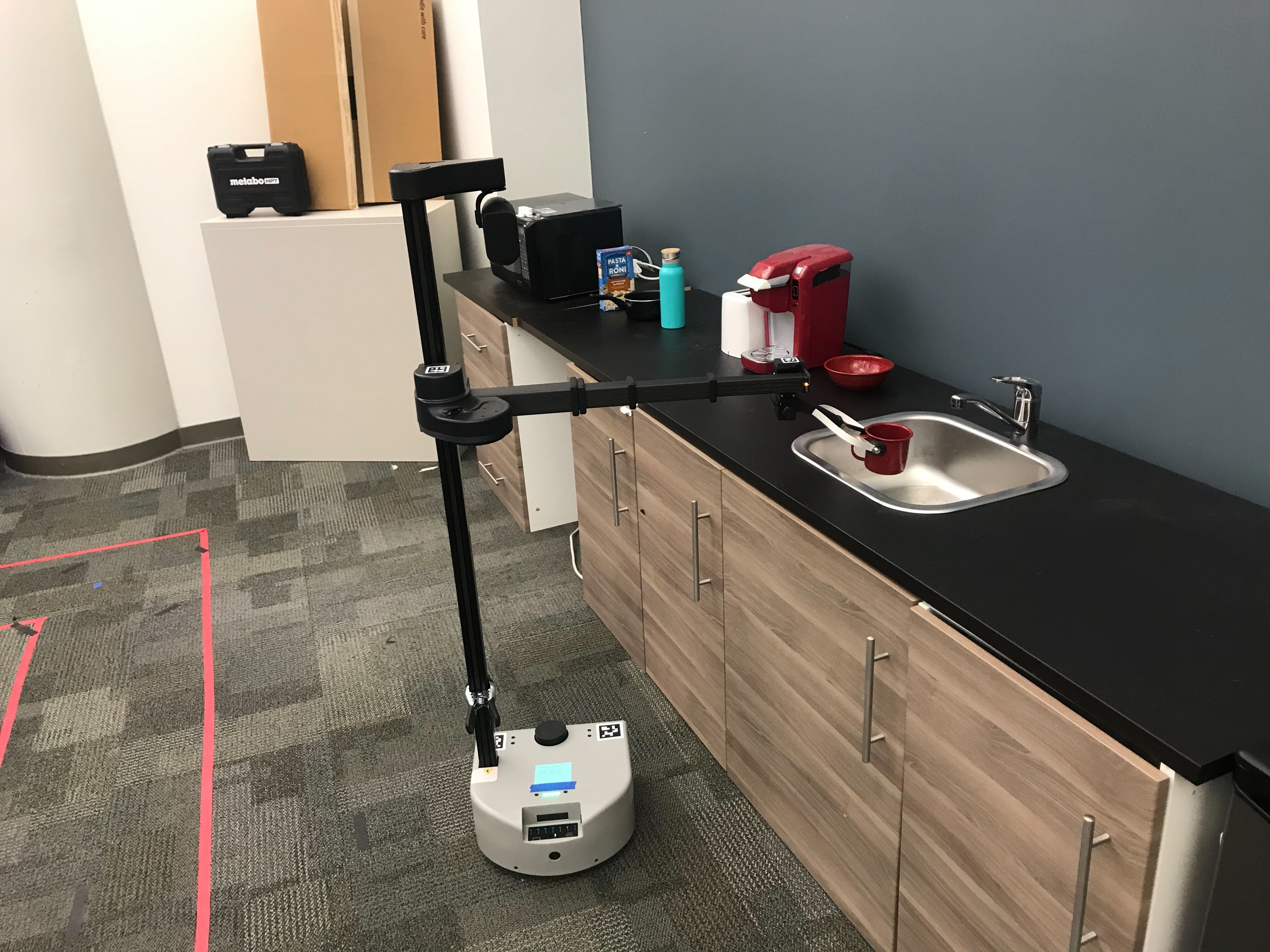}
         \caption{Place mug}
     \end{subfigure}
     \begin{subfigure}{0.18\textwidth}
         \centering
         \includegraphics[width=\textwidth]{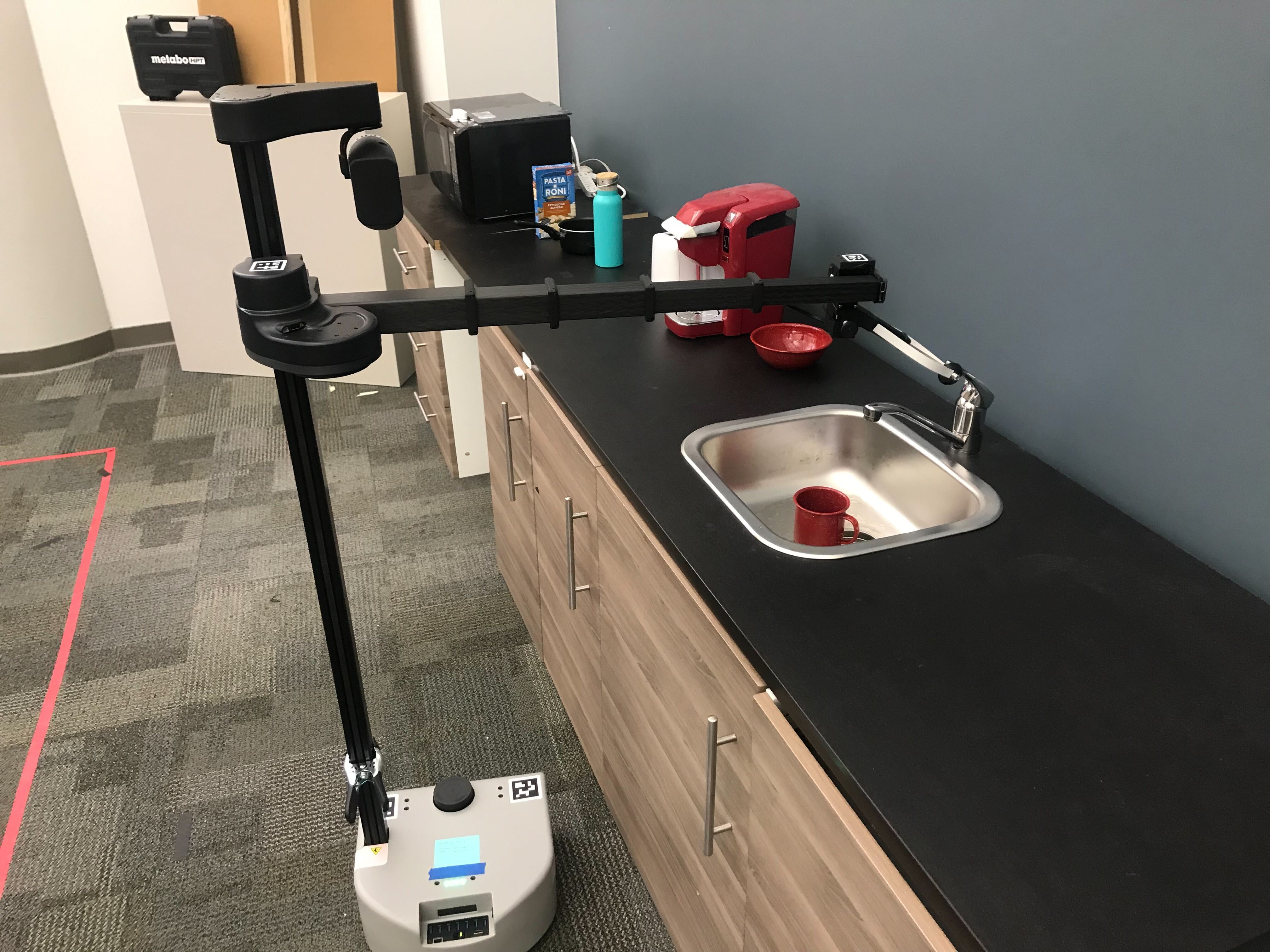}
         \caption{Toggle faucet}
     \end{subfigure}
     \caption{Real world human instruction following (clean test).}
     \label{fig:hif_clean_test}
\end{figure*}

\textbf{Object Navigation}: The criterion of a successful object search requires the robot to search at most two target areas (relative to a selected stationary object) to find the target object. The targets objects include $\{\texttt{bottle},\texttt{knife},\texttt{plate}, \texttt{pot},\texttt{toaster}, \texttt{mug},\texttt{sponge}, \\ \texttt{bowl}, \texttt{book},\texttt{banana}\}$. We conduct 10 evaluation trials, with a success rate of 6 out of 10. See \cref{fig:success real world object search} for a successful search example. The failed trials including searching for $\{\texttt{book}, \texttt{banana},\texttt{bowl}, \texttt{knife}\}$. The primary cause for failure is the bias of the GNN model inherited from the Visual Genome training data. Specifically, the model often predicts \texttt{sink} or \texttt{counter} as related objects, as the co-occurrence between \texttt{sink} or \texttt{counter} and other target objects are common in the dataset. Although GNN detects objects and updates information during the search, it may still converge to incorrect predictions in some cases. For instance, when searching for \texttt{banana}, the model initially predicts \texttt{sink}, followed by \texttt{counter} and \texttt{desk} and misses the correct prediction \texttt{table}.

\textbf{Human Instruction Following}: Due to mechanical limitation of the gripper, we limit our task to the ones that only involve transporting objects and simple kinematic motion. 
\begin{itemize}
\item \textbf{Pick-and-place.} \emph{Targets} are \texttt{mug}, \texttt{bottle}, \texttt{orange}, \texttt{box}; Based on human instruction, the robot should correctly plan and execute a sequence of actions including searching for and picking up the \emph{target} and placing it on the \texttt{counter} or \texttt{desk}.
\item \textbf{Clean.} \emph{Target} is \texttt{mug}; Based on human instruction, the robot should correctly plan and execute a sequence of actions including searching for and picking up the \emph{mug}, placing it inside \texttt{sink}, and toggling to turn on the \texttt{faucet}.
\end{itemize}

For each pick-and-place task, we place one target object on the table. The evaluation success rate is 3 out of 4 objects. See \cref{fig:hif_pp_test} for the successful cases. The failed trial is caused by the perception module failing to estimate the correct 3D position for the \emph{orange}, which results in a failed grasp. In the clean task, we place the mug on the table and change its location slightly each time, repeating the experiment four times. The success rate is 3 out of 4 times, see \cref{fig:hif_clean_test} for visualization. The robot is able to locate the mug reliably. The failure case is again caused by a failed grasp (the mug slipped out of the gripper).

\section{Conclusion}
This work proposes a zero-shot object navigation approach, which leverages object co-occurrence to achieve efficient search. Unlike prior correlation-based methods, we learn the object correlation from a large-scale Visual Genome dataset and don't require accurate correlation information in new environments. We apply this approach to object navigation and integrate it into human instruction following tasks, and demonstrate the effectiveness in both AI2-THOR simulator and real-world household settings. The current limitations of our system lie in the object manipulation and grasping algorithms. For instance, the system fails to open containers for object detection and struggles to hold onto objects securely. Exploring more advanced planners that can take additional manipulation actions to search object and learning to predict viable grasp candidates~\cite{chu2018real} would be interesting for future works.

{\small
\bibliographystyle{unsrt}
\bibliography{root}
}

\end{document}